\documentclass{article}

\usepackage{microtype}
\usepackage{graphicx}
\usepackage{subcaption}
\usepackage{booktabs} 

\usepackage{hyperref}

\usepackage[table]{xcolor}
\usepackage{multirow}
\usepackage{tabularx}
\usepackage{amsmath}
\usepackage{algorithmic}
\usepackage{array}
\newcolumntype{Y}{>{\centering\arraybackslash}X}
\usepackage{makecell} 

\usepackage[accepted]{icml2025}

\usepackage{amssymb}
\usepackage{mathtools}
\usepackage{amsthm}
\usepackage{fontawesome5}

\usepackage[capitalize,noabbrev]{cleveref}

\theoremstyle{plain}

\theoremstyle{definition}

\theoremstyle{remark}

\usepackage[textsize=tiny]{todonotes}

\icmltitlerunning{Self-Correcting VLA: Online Action Refinement via Sparse World Imagination}

\begin{document}

\twocolumn[
\icmltitle{Self-Correcting VLA: Online Action Refinement via Sparse World Imagination}



\icmlsetsymbol{equal}{*}
\icmlsetsymbol{corr}{\faEnvelope}

\begin{icmlauthorlist}
\icmlauthor{Chenyv Liu}{Tongji,equal}
\icmlauthor{Wentao Tan}{Tongji,equal}
\icmlauthor{Lei Zhu}{Tongji,corr}
\icmlauthor{Fengling Li}{UTS}
\icmlauthor{Jingjing Li}{UESTC}
\icmlauthor{Guoli Yang}{AIBD}
\icmlauthor{Heng Tao Shen}{Tongji}
\end{icmlauthorlist}

\icmlaffiliation{Tongji}{Tongji University}
\icmlaffiliation{UTS}{University of Technology Sydney}
\icmlaffiliation{UESTC}{University of Electronic Science and Technology of China}
\icmlaffiliation{AIBD}{Advanced Institute of Big Data}

\icmlcorrespondingauthor{Lei Zhu}{leizhu0608@gmail.com}

\icmlkeywords{Machine Learning, Robot Learning}

\vskip 0.3in
]

\begin{figure*}[!ht]
    \centering
    \includegraphics[width=1\textwidth]{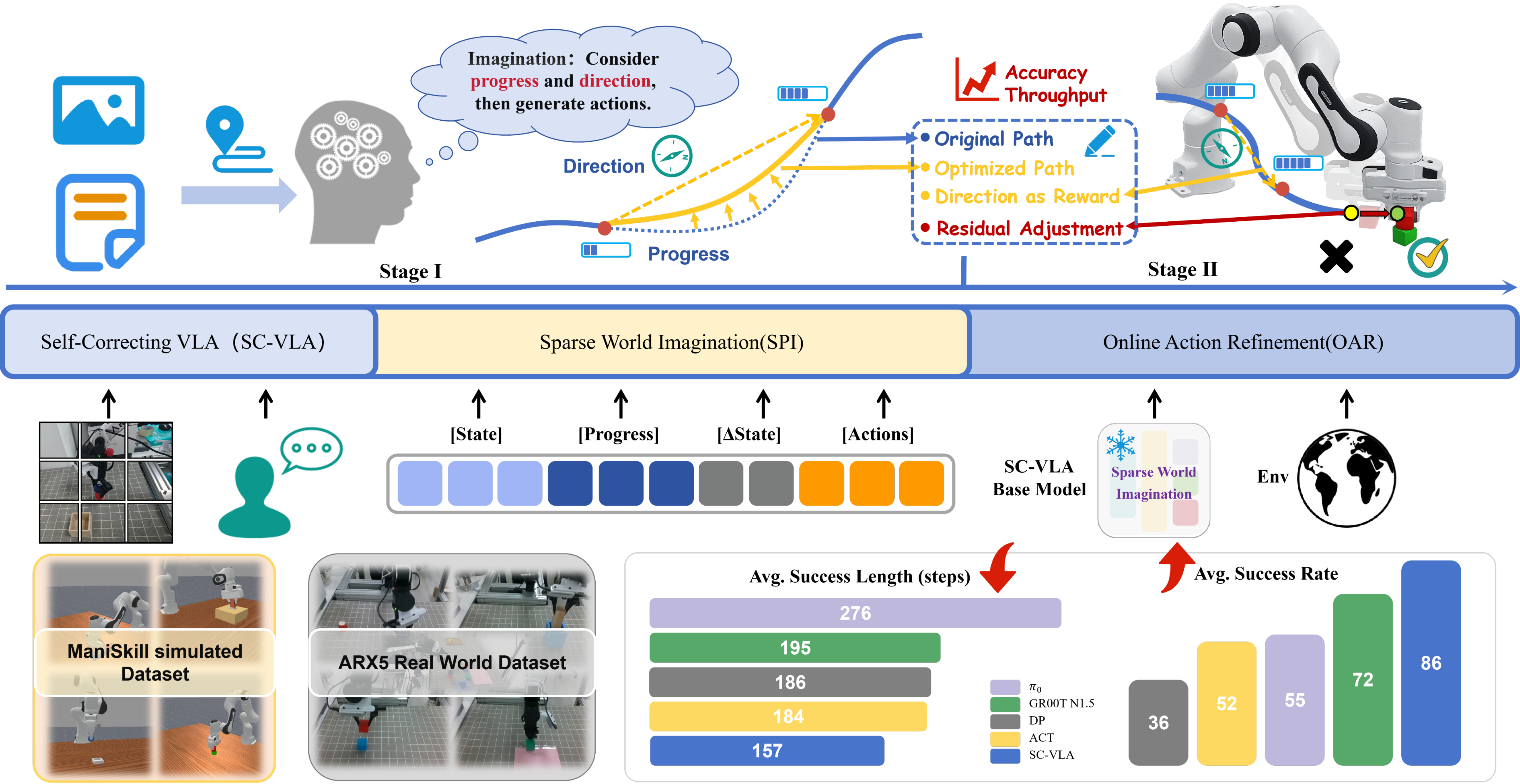}
    \caption{We present Self-Correcting VLA (SC-VLA), a novel framework designed to enhance physical grounding through intrinsic self-improvement. The model is equipped with Sparse World Imagination (SPI) to forecast task progress and future trajectory trends, and Online Action Refinement (OAR) to dynamically optimize policies via residual adjustments and reshaped rewards. SC-VLA achieves superior performance on ManiSkill and real-world ARX5 benchmarks, surpassing baselines in both success rate and execution throughput.}
    \label{fig:sc_vla_general_framework}
    \vspace{-0.2cm}
\end{figure*}


\printAffiliationsAndNotice{\icmlEqualContribution} 

\begin{abstract} 
Standard vision-language-action (VLA) models rely on fitting statistical data priors, limiting their robust understanding of underlying physical dynamics. Reinforcement learning enhances physical grounding through exploration yet typically relies on external reward signals that remain isolated from the agent's internal states. World action models have emerged as a promising paradigm that integrates imagination and control to enable predictive planning. However, they rely on implicit context modeling, lacking explicit mechanisms for self-improvement. To solve these problems, we propose Self-Correcting VLA (SC-VLA), which achieve self-improvement by intrinsically guiding action refinement through sparse imagination. We first design sparse world imagination by integrating auxiliary predictive heads to forecast current task progress and future trajectory trends, thereby constraining the policy to encode short-term physical evolution. Then we introduce the online action refinement module to reshape progress-dependent dense rewards, adjusting trajectory orientation based on the predicted sparse future states. Evaluations on challenging robot manipulation tasks from simulation benchmarks and real-world settings demonstrate that SC-VLA achieve state-of-the-art performance, yielding the highest task throughput with \textbf{16\%} fewer steps and a \textbf{9\%} higher success rate than the best-performing baselines, alongside a \textbf{14\%} gain in real-world experiments. Code is available at \url{https://github.com/Kisaragi0/SC-VLA}.
\end{abstract}

\section{Introduction} 
Vision-language-action models (VLA) have driven recent advancements in embodied AI by bringing multimodal large language models (MLLMs) to physical control~\cite{rt-2,palm-e,openvla,octo}. By performing large-scale imitation learning on diverse robotic datasets, these architectures enable agents to translate natural language instructions directly into executable actions. However, this paradigm fundamentally relies on fitting the statistical patterns inherent in the pre-training data. Consequently, the learned policies mainly depend on memorized data priors rather than acquiring a robust understanding of the underlying physical dynamics~\cite{physbench,tracevla}.

To address the reliance on static priors, reinforcement learning introduces active interaction into the training process. Instead of passively cloning expert behaviors, agents optimize their policies by exploring the environment and learning from rewards. 
To implement this, several research paradigms have emerged. A prominent strategy employs online reinforcement learning algorithms, such as PPO~\cite{ppo}, to fine-tune pre-trained VLA for task-specific adaptation~\cite{can-rl-bring,vla-rl}. Alternatively, offline reinforcement learning methods seek to extract optimal policies from static datasets, enhancing performance without necessitating real-time interaction~\cite{conservative,q-transformer}. However, effective learning relies on the quality of feedback signals, which are often difficult to define manually for diverse tasks and scenarios. To address this, recent methods incorporate multimodal large language models to synthesize reward functions, using semantic reasoning to guide policy optimization~\cite{vla-rl,vlac}.
Despite these advancements, a critical limitation persists: \textit{whether manually defined or synthesized by models, these approaches typically rely on external reward signals to evaluate performance.} This reliance can introduce a disconnect between the external signals and the model's internal states. Therefore, it is essential to explore efficient self-improvement strategies that facilitate the model's native adaptation capabilities.

To bridge this gap, world models offer a promising direction by establishing internal physical dynamics. However, current reinforcement learning algorithms for VLA typically treat world models and policies as independent modules, relying on closed-loop feedback for exploration~\cite{nora}. This separation overlooks the unique capability of another paradigm, termed \textbf{world action models}~\cite{gr-1,cot-vla,susie}, which natively support both action generation and multimodal future prediction. While this allows intrinsic evaluation via predicted future states, existing approaches lack explicit mechanisms to leverage these states to refine actions, failing to realize self-improvement aligned with the agent's internal states.

To achieve self-improvement by intrinsically guiding action refinement through future imagination, we propose Self-Correcting VLA (SC-VLA). This framework jointly generates actions while predicting sparse future states, enabling fine-grained trajectory refinement via residual reinforcement learning for complex manipulation tasks. SC-VLA significantly enhances task success rate and throughput. Specifically, we introduce sparse world imagination, which integrates auxiliary predictive heads to forecast current task progress and future trajectory orientation as sparse world signals. This mechanism constrains the policy to encode short-term physical evolution prior to action generation. Based on this, we further propose online action refinement, which constructs dense rewards by evaluating the consistency between the current and future trajectory orientation. It employs imagination to deduce future action trends for reshaping intrinsic progress-dependent dense rewards, thereby eliminating reliance on external reward models. We evaluate our method on four challenging manipulation tasks (StackCube, PlaceSphere, LiftPegUpright, and PegInsertion), where SC-VLA demonstrates state-of-the-art performance with the highest task throughput.
\begin{itemize}
    \item We propose SC-VLA, a self-correcting framework that integrates offline action generation with online refinement. By introducing sparse world imagination to forecast sparse future states, it constrains the policy to encode physical evolution.
    \item We develop online action refinement with residual reinforcement learning to adjust trajectory orientation. It constructs progress-dependent dense rewards using predicted future states, explicitly guiding the policy to align with the imagined future behavioral trends.
    \item We conduct systematic evaluations on four tasks across both simulation and real-world robotic platform. The results demonstrate that SC-VLA achieves the best performance in task throughput and success rate on complex manipulation tasks.
\end{itemize}

\section{Related Works}
\paragraph{Vision-Language-Action Models.}
Vision-Language-Action (VLA) models integrate multimodal models into robotic control~\cite{rt-1,roboflamingo,rdt}. RT-2~\cite{rt-2} enables semantic transfer by modeling actions as discrete tokens, while OpenVLA~\cite{openvla} enhances cross-robot generalization through efficient adaptation. Recently, GR00T~\cite{gr00t} and $\pi_0$~\cite{pi_0} have improved real-time deployment and continuous action generation via dual-system architectures and flow matching, respectively. Despite these advances, existing methods rely heavily on offline imitation and current-step semantic alignment, lacking explicit modeling of physical evolution.

\paragraph{Reinforcement Learning for VLA.}
The key to introducing reinforcement learning into pretrained VLA models lies in constructing effective external rewards to alleviate the sparse feedback problem. Existing approaches can be broadly categorized into three classes. The first class leverages the semantic reasoning capability of vision-language models (VLM) to directly evaluate execution processes or diagnose failure causes, thereby synthesizing trajectory-level rewards or corrective signals, such as VLA-RL~\cite{vla-rl}, VLAC~\cite{vlac}, Reflective Self-Adaptation~\cite{reflection}, and World-Env~\cite{world-env}, which provide high-level semantic guidance through language understanding and causal reasoning. The second class designs rewards via explicit rules or auxiliary prediction objectives, including rule-based signals derived from remaining steps or temporal information (e.g., Self-Improving~\cite{self-improving}, $\pi^{*}_{0.6}$~\cite{pi_0.6}, and RFTF~\cite{rftf}), as well as approaches such as NORA-1.5~\cite{nora} that evaluate trajectory deviations using action-conditioned world models. The third class constructs rewards based on feature or trajectory similarity metrics, typically by comparing generated trajectories with target trajectories in pixel, perceptual, or latent feature spaces, as exemplified by ThinkAct~\cite{thinkact} and VLA-RFT~\cite{vla-rft}. Despite significantly improving exploration efficiency, these methods generally rely on external model inference, handcrafted rules, or similarity computations, which are decoupled from the policy's internal representations and introduce additional computational and system complexity. In contrast, approaches that rely solely on scalar return prediction (e.g., ReinboT~\cite{reinbot}) struggle to provide fine-grained spatiotemporal geometric constraints.

\paragraph{World Action Models.}
Unlike RL-based VLA methods that rely on external reward signals, world action models aim to jointly model action generation and future evolution within a unified framework, constraining policy behavior via latent contextual predictions and thereby enabling intrinsic guidance without explicit external rewards. GR-MG~\cite{grmg} conditions current actions on imagined future target images by generating pixel-level goals. FLARE~\cite{flare} further avoids high-dimensional pixel synthesis by aligning future representations in latent space, improving both efficiency and stability. PAR~\cite{par} proposes a physics-aware autoregressive modeling paradigm that unifies vision and actions as continuous physical tokens, directly leveraging dynamical priors from video models for control. WorldVLA~\cite{worldvla} alternates between predicting actions and future visual states in an autoregressive manner, mitigating error accumulation through attention masking and achieving mutual reinforcement between action generation and world modeling. Despite their ability to constrain policies with predictive latent context, the future signals in these approaches are typically encoded as implicit representations, lacking interpretable physical semantics or explicit self-evaluation mechanisms. As a result, they struggle to perform fine-grained corrections on short-horizon trajectories and cannot provide direct policy improvement signals in the same manner as reward-based paradigms.

\section{Preliminary}
\subsection{Basic Robot Policy}
To address the multimodal distribution inherent in robot action generation, we adopt Flow Matching (FM) as the backbone of our policy. Compared to diffusion-based models, flow matching constructs deterministic optimal transport paths, significantly improving training efficiency and inference stability while preserving generation quality. 

Concretely, given an observation $o$, conditional flow matching aims to learn an observation-conditioned vector field $v_\theta$ that continuously transforms samples from a prior noise distribution $p_0(x) = \mathcal{N}(x \mid 0, I)$ into a target action distribution $p_1(x) \approx q(a \mid o)$. During training, we construct an optimal transport interpolation path
\begin{equation}
\\x_t = t x_1 + (1 - t) x_0,
\end{equation}
which connects a noise sample $x_0 \sim p_0$ and a ground-truth action sample $x_1 \sim q(a \mid o)$. The corresponding target velocity field is given by $x_1 - x_0$. The model parameters are optimized by minimizing the mean squared error between the predicted vector field and the target velocity field:
\begin{equation}
\mathcal{L}_{\text{FM}}(\theta)
=
\mathbb{E}_{t, x_0, x_1, o}
\left[
\left\|
v_\theta\big(\\x_t, t, o\big)
-
(x_1 - x_0)
\right\|_2^2
\right].
\end{equation}

At inference time, the model starts from a prior noise sample and generates the final action by solving the corresponding ordinary differential equation (ODE) with a numerical integrator (e.g., Euler stepping):
\begin{equation}
x_1 = x_0 + \int_{0}^{1} v_\theta(x_t, t, o)\, dt.
\end{equation}
In the following sections, we build upon this conditional flow matching formulation and introduce additional structured predictive constraints to enhance the policy's ability to model short-term physical evolution.

\subsection{Soft Actor-Critic (SAC)}
To enable stable and efficient policy optimization in continuous action spaces, we adopt Soft Actor-Critic (SAC)~\cite{sac} as the underlying reinforcement learning algorithm. SAC is an off-policy method based on the maximum entropy reinforcement learning framework, which augments the expected return objective with an entropy regularization term. This formulation encourages a balance between exploration and exploitation, leading to improved stability and robustness of the learned policy. The optimization objective of SAC is defined as:
\begin{equation}
J(\pi)
=
\mathbb{E}_{\tau \sim \pi}
\left[
\sum_{t=0}^{\infty}
\gamma^t
\left(
r(s_t, a_t)
+
\alpha \, \mathcal{H}\big(\pi(\cdot \mid s_t)\big)
\right)
\right],
\end{equation}
where $\mathcal{H}(\pi(\cdot \mid s_t))$ denotes the entropy of the policy at state $s_t$, and $\alpha$ is an automatically tuned temperature parameter that controls the strength of entropy regularization.

In continuous action spaces, SAC typically updates the policy using the reparameterization trick, in which action sampling is expressed as
\begin{equation}
a_t = f_\theta(s_t, \xi), \qquad \xi \sim \mathcal{N}(0, I),
\end{equation}
enabling low-variance gradient backpropagation through the stochastic policy.

\section{Self-Correcting VLA}
\begin{figure*}[t]
    \centering
    \includegraphics[width=0.95\textwidth]{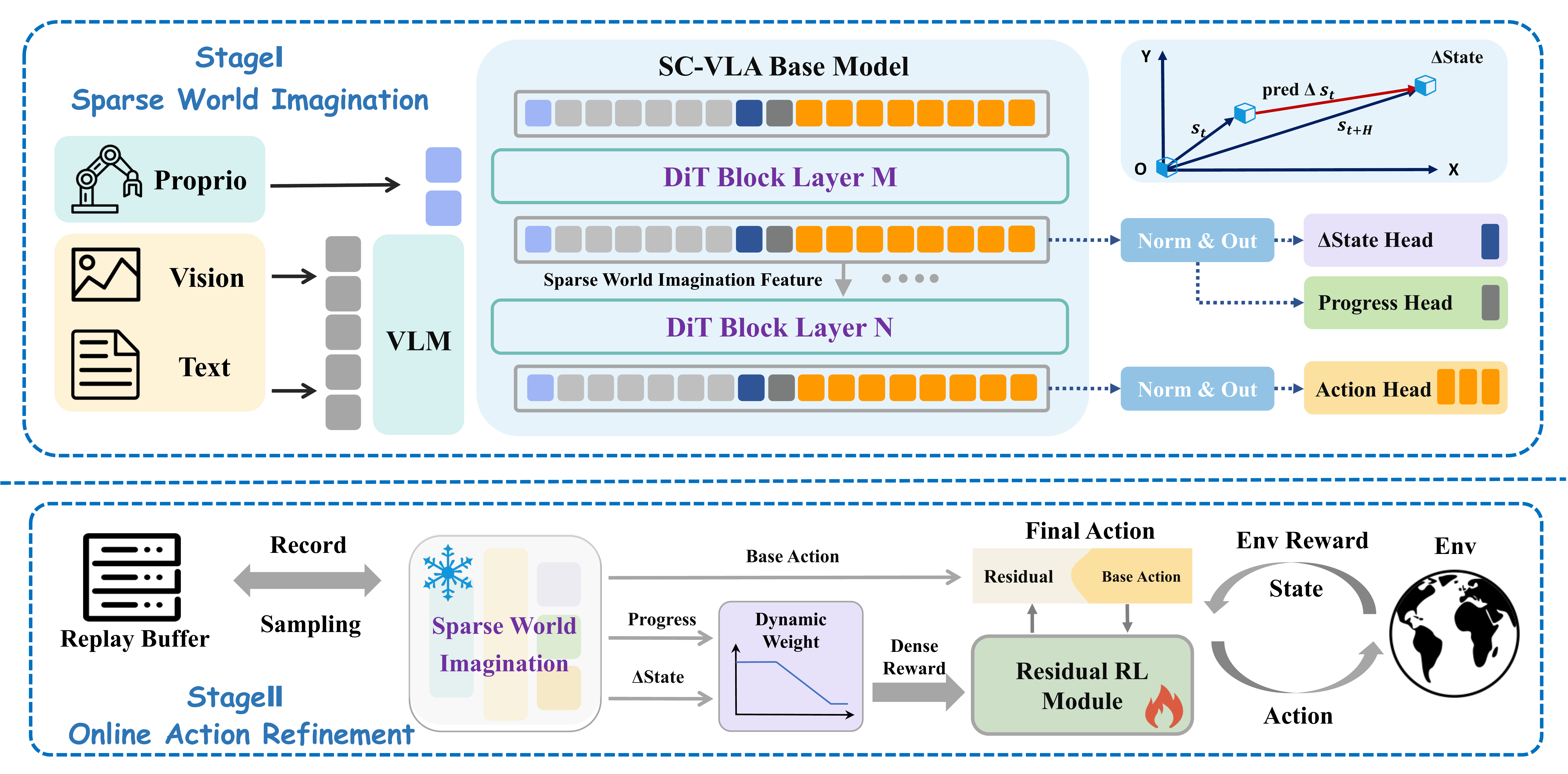}
    \caption{The architecture of Self-Correcting VLA. The framework consists of two stages: Stage I (Top) utilizes a VLM and DiT-based backbone to generate base actions and sparse world imagination, decoded from the final output (Layer $N$) and intermediate features (Layer $M$), respectively. Stage II (Bottom) implements Online Action Refinement, where a Residual RL Module optimizes the final action by learning a residual term. This process is guided by endogenous dense rewards derived from the dynamic weighting of imagination consistency (Progress and $\Delta$State ) without external supervision.}
    \label{fig:sc_vla_detail_framework}
    \vspace{-0.2cm}
\end{figure*}

We present Self-Correcting VLA (SC-VLA), a two-stage framework illustrated in Fig.~\ref{fig:sc_vla_detail_framework}. SC-VLA integrates sparse world imagination (SPI) with residual reinforcement learning for online action refinement (OAR). Crucially, the reward signals are \textbf{endogenous}, derived purely from the internal consistency of world imagination rather than external dense supervision. This design comprises two key components: (i) a flow-matching base policy guided by sparse imagination (Sec.~4.1), and (ii) a residual module optimized via intrinsic predictive rewards (Sec.~4.2).

\subsection{Sparse World Imagination}
\paragraph{Conditional Information Processing.}
We leverage a Vision-Language Model (VLM) to fuse multimodal observations into a unified representation. Specifically, multi-view images $I_k$ are encoded by the SigLIP-2~\cite{siglip} backbone and concatenated with the language instruction $\mathcal{L}$. This joint sequence is then processed by Eagle-2~\cite{eagle}. To balance high-level semantic understanding with low-level control features, we extract the hidden states from an intermediate layer $l$ as the conditioning signal $\mathbf{o}_{\mathrm{mid}}$:
\begin{equation}
    \mathbf{o}_{\mathrm{mid}} = \Phi_{\mathrm{VLM}}^{(l)} \big( \mathcal{E}_{\mathrm{vis}}(I_k), \mathcal{L} \big),
\end{equation}
where $\mathcal{E}_{\mathrm{vis}}$ denotes the SigLIP-2 encoder and $\Phi_{\mathrm{VLM}}^{(l)}$ represents the VLM backbone truncated at layer $l$.

\paragraph{Query Sequence Construction.}
To compensate for the lack of physical consistency constraints in standard flow matching, we introduce a physics regularization mechanism based on sparse world imagination. Explicit world prediction targets are injected into the query sequence to be jointly modeled with the action vector field. The augmented input query sequence is defined as:
\begin{equation}
\mathbf{q}_{\mathrm{input}} = [\mathbf{s}_t, \mathbf{q}_{p_t}, \mathbf{q}_{\Delta s_t}, \mathbf{q}_a ].
\end{equation}
Here, $\mathbf{s}_t \in \mathbb{R}^{1 \times D}$ denotes the current embodiment state, and $\mathbf{q}_a \in \mathbb{R}^{16 \times D}$ corresponds to the queries for action trajectory generation. These two components serve as the base queries for action generation.
$\mathbf{q}_{p_t} \in \mathbb{R}^{1 \times D}$ is used to predict the task progress $p_t$, providing the model with explicit temporal evolution cues.
$\mathbf{q}_{\Delta s_t} \in \mathbb{R}^{1 \times D}$ is introduced to model short-horizon future physical state variations, capturing the spatial displacement trends induced by the action.
Specifically, we model short-term physical evolution as a relative transformation within the current local coordinate frame. We define the target state at a future time step $t' = t + H + \delta$, where $H$ is the execution horizon and $\delta \sim \mathcal{U}(-\Delta, \Delta)$ is a random temporal offset introduced for robustness. The relative state increment $\Delta s_t \in \mathbb{R}^7$ is calculated as:
\begin{equation}
    \Delta s_t =
    \big[
    R_t^\top (P_{t'} - P_t),\;
    \mathrm{Euler}(R_t^\top R_{t'} ),\;
    g_{t'} - g_t
    \big].
\end{equation}
Here, $P \in \mathbb{R}^3$, $R \in SO(3)$, and $g \in \mathbb{R}$ denote the end-effector position, rotation matrix, and gripper opening, respectively. The operator $\mathrm{Euler}(\cdot)$ extracts Euler angles from the relative rotation matrix. This sparse, local-frame prediction objective enhances the model's generalization to physical dynamics across varying temporal scales.

\paragraph{Joint Optimization.}
In the DiT backbone composed of $N$ Transformer blocks, the final block primarily focuses on modeling the action distribution, while intermediate layers retain explicit representations of the world state. To encode short-term physical evolution prior to action generation, we extract the hidden representation $h^{(m)}$ from the $m$-th intermediate block. Based on this feature, we predict the task progress $p_t$ and relative state change $\Delta s_t$ using independent lightweight MLP heads:
\begin{equation}
\hat{p}_t = f_{\mathrm{prog}}(h^{(m)}_{p_t}), 
\qquad 
\widehat{\Delta s}_t = f_{\Delta s_t}(h^{(m)}_{\Delta {s_t}}).
\end{equation}
We jointly optimize the flow matching objective with this auxiliary physical supervision. The overall training objective is defined as:
\begin{equation}
\mathcal{L}_{\mathrm{total}} 
= 
\mathcal{L}_{\mathrm{FM}} 
+ 
\lambda_1 \mathcal{L}_{\mathrm{prog}} 
+ 
\lambda_2 \mathcal{L}_{\Delta s_t},
\end{equation}
where $\mathcal{L}_{\mathrm{prog}}$ and $\mathcal{L}_{\Delta s_t}$ are supervised using Mean Squared Error (MSE), weighted by coefficients $\lambda_1$ and $\lambda_2$. Through this joint optimization, the model not only learns to generate precise actions but also internalizes a coherent and interpretable sparse world representation within $h^{(m)}$, providing robust guidance for the subsequent residual policy adaptation.

\subsection{Online Action Refinement}
While the base policy (Sec.~4.1) enhances stability by encoding physical evolution, it remains limited by offline data, often struggling with out-of-distribution perturbations and fine-grained contacts. To address this without learning from scratch, we introduce a residual RL module atop the base priors. This module performs minimal online corrections to base actions, enabling effective adaptation in high-precision settings.

\paragraph{Residual Policy.}
To balance the stability of the base policy and online adaptability to environmental perturbations, we adopt a residual policy structure. The input space of the residual policy $\pi_{\mathrm{res}}$ is reconstructed as a \emph{sparse world imagination observation} $o_w \in \mathbb{R}^{16}$:
\begin{equation}
\label{eq:o_w}
o_w = (s_t, \hat{p}_t, \widehat{\Delta s}_t).
\end{equation}
In this work, we follow the Policy Decorator~\cite{policy-decorator} to model the residual policy $\pi_{\mathrm{res}}$ as a Gaussian policy parameterized by a lightweight multilayer perceptron (MLP) and adopt the Soft Actor-Critic (SAC)~\cite{sac} algorithm  because it has excellent sample efficiency and stability.
The final action $a_t$ is jointly determined by a frozen base policy $a_t^{\mathrm{base}} \sim \pi_{\mathrm{base}}(o_{\mathrm{mid}})$ and a learnable residual policy $a_t^{\mathrm{res}} \sim \pi_{\mathrm{res}}(o_w)$:
\begin{equation}
a_t = a_t^{\mathrm{base}} + \lambda a_t^{\mathrm{res}},
\end{equation}
where $\lambda$ is a residual scaling coefficient. By explicitly incorporating task progress and state evolution predicted by the first-stage model, the residual network can perceive the intent of the base policy. This design enables the residual policy to perform local adjustments around the physical evolution priors provided by the model, thereby avoiding inefficient and unconstrained exploration in the raw observation space.

\paragraph{Dense Reward Mechanism.}
Although the residual architecture reduces the dimensionality of exploration, the environment reward remains highly sparse. To address this issue, we construct a directional dense guidance reward using the short-term state change $\widehat{\Delta s}_t$ predicted by the base policy. Specifically, we extract only the first three translational components $\Delta s_t^{\mathrm{pos}} \in \mathbb{R}^3$ from $\Delta s_t$ and define the short-term goal position at the current timestep as
\begin{equation}
P_{\mathrm{goal}} = P_t + \widehat{\Delta s}_t^{\mathrm{pos}}.
\end{equation}
After executing the residual action, the guidance reward is computed based on the alignment between the actual end-effector displacement and the predicted evolution direction:
\begin{equation}
r_t^{\mathrm{guide}} =
\frac{(P_{t+n} - P_t) \cdot (P_{\mathrm{goal}} - P_t)}
{\lVert P_{t+n} - P_t \rVert \, \lVert P_{\mathrm{goal}} - P_t \rVert + \epsilon}.
\end{equation}
Here, $P_{t+n}$ denotes the end-effector position after executing the action for $n$ steps $(n < H)$. This reward provides continuous directional feedback at each step, guiding the residual policy to perform local adjustments along the short-term physical evolution direction predicted by the base policy, thereby effectively alleviating exploration difficulties under sparse reward settings.

\begin{algorithm}[h]
\caption{SC-VLA: Online Action Refinement}
\label{alg:mpg_rl}
\begin{algorithmic}[1]
\STATE \textbf{Input:} Frozen SC-VLA base policy $\pi_{\text{base}}$, online action refinement policy $\pi_{\text{res}}$, environment Env
\STATE Initialize replay buffer $\mathcal{D}$
\STATE Initialize total training steps $T_{\text{train}}$
\STATE $t \leftarrow 0$
\WHILE{$t < T_{\text{train}}$}
    \STATE Observe state $s_t$
    \STATE Query base policy:
    $(a_t^{\text{base}},\, \hat{p}_t,\, \widehat{\Delta s}_t) \sim \pi_{\text{base}}(o_\text{mid})$
    \STATE Construct augmented observation $o_{w}^{t}$ by Eq.~\eqref{eq:o_w}
    \STATE Sample residual action $a_t^{\text{res}} \sim \pi_{\text{res}}(o_{w}^{t})$
    \STATE Execute action $a_t \leftarrow a_t^{\text{base}} + \lambda a_t^{\text{res}}$
    \STATE Step environment:
    $(o_{w}^{t+1}, r_t^{\text{env}}) \leftarrow \text{Env.Step}(a_t)$
    \STATE Compute final reward  $r_{t}^{final}$ by Eq.~\eqref{eq:final_reward}
    \STATE Add $(o_{w}^{t}, a_t^{\text{base}}, a_t^{\text{res}} , o_{w}^{t+1}, a_{t+1}^{\text{base}},r_t^{final})$ in $\mathcal{D}$
    \STATE Update $\pi_{\text{res}}$ and critics using SAC
    \STATE $t \leftarrow t + 1$
\ENDWHILE
\end{algorithmic}
\end{algorithm}

\paragraph{Dynamic Weight Scheduling.}
While sparse world imagination accelerates global exploration, a static predictive prior may limit the policy's optimization capability during fine-grained contact phases due to distribution shift. To address this issue, we propose a \emph{dynamic weight scheduling}. Specifically, we use the predicted task progress $\hat{p}_t$ as a scheduling signal, allowing predictive guidance to dominate in the early stages of the task and to be gradually weakened in later stages, thereby enabling a smooth transition from predictive-prior guidance to autonomous exploration. The final reward is defined as a weighted combination of the environment reward and the predictive guidance reward:
\begin{equation}
\label{eq:final_reward}
r_t^{\mathrm{final}}
=
\eta(\hat{p}_t) \cdot w_{\mathrm{guide}} \cdot r_t^{\mathrm{guide}}
+
r_t^{\mathrm{env}}
-
c,
\end{equation}
where $\eta(\cdot)$ is a monotonically decreasing scheduling function with respect to task progress, $w_{\mathrm{guide}}$ is the guidance weight, $r_t^{\mathrm{env}}$ denotes the sparse environment reward, and $c$ is a per-step time penalty. This mechanism effectively balances early exploration efficiency and late-stage sensitivity to real dynamics feedback, ensuring stable convergence of the policy during the fine-tuning phase.

\section{Experiments}
In this section, we systematically evaluate the effectiveness of \emph{Self-Correcting Vision-Language-Action} (SC-VLA) through both simulation and real-robot experiments, with a primary focus on success rate, throughput and transferability in complex manipulation tasks. Specifically, our experimental analysis is organized around the following key questions:
\begin{enumerate}
    \item Can SC-VLA improve the success rate of flow-matching policies in complex manipulation tasks through sparse world imagination combined with a residual module?
    \item Can the dense reward constructed based on sparse world imagination and the dynamic weight scheduling alleviate the issue of low exploration efficiency caused by sparse rewards and improve policy throughput?
    \item What are the respective contributions of the key components in the proposed framework to the overall performance?
    \item Can the proposed method be stably transferred to real robotic systems and maintain robustness under environmental perturbations?
\end{enumerate}

\subsection{Simulation Setup and Baselines}
\label{sec:sim_setup}
\paragraph{ManiSkill3.}
We evaluate our method on ManiSkill3~\cite{maniskill3}, a SAPIEN-based~\cite{sapien} platform offering high-fidelity contact dynamics suitable for complex manipulation. We select four challenging tasks: \emph{StackCube}, \emph{PlaceSphere}, \emph{PegInsertion}, and \emph{LiftPegUpright}, to cover capabilities ranging from precise pick-and-place to non-prehensile manipulation. For fair comparison, all methods are trained with 100 demonstrations per task and evaluated over 50 episodes. See Appendix A for details.

\paragraph{Baselines.}
We compare SC-VLA against baselines representing mainstream paradigms: \emph{Diffusion Policy} (DP)~\cite{diffusion} for diffusion-based control, \emph{ACT}~\cite{act} for action chunking, and $\pi_0$~\cite{pi_0} for flow-matching policies, alongside our base model GR00T N1.5~\cite{gr00t}. We use task success rate as the primary metric. Additionally, we report average episode length during the RL stage to evaluate exploration efficiency and system throughput. Detailed configurations are provided in Appendix B.

\begin{table}[htbp] 
    \centering
    \small
    \setlength{\tabcolsep}{5pt} 
    \renewcommand{\arraystretch}{1.15}

    \definecolor{mpgRed}{RGB}{255,235,235}
    \definecolor{sacBlue}{RGB}{235,240,255}
    \definecolor{deltaGreen}{RGB}{0,150,0}
    \definecolor{avgGray}{RGB}{245,245,245}

    \caption{Success rates of all methods in ManiSkill. Due to the absence of language guidance in DP, ACT, we evaluate them under two settings: $^{\ddagger}$ denotes a single multi-task policy trained on all tasks simultaneously. $^{\dagger}$ denotes training independent specialist models for each of the four tasks separately.}
    \label{tab:success_rates}

    \resizebox{\linewidth}{!}{%
        \begin{tabular}{lcccc>{\columncolor{avgGray}}c}
            \toprule
            \multirow{2}{*}{\textbf{Model}} &
            \multicolumn{4}{c}{\textbf{ManiSkill Tasks}} &
            \multicolumn{1}{c}{\multirow{2}{*}{\textbf{Avg}}} \\
            \cmidrule(lr){2-5}
            & \textbf{\makecell{Stack\\Cube}} 
            & \textbf{\makecell{Place\\Sphere}} 
            & \textbf{\makecell{LiftPeg\\Upright}} 
            & \textbf{\makecell{Peg\\Insertion}}
            & \multicolumn{1}{c}{} \\
            \midrule
      
            DP$^{\ddagger}$ & 0.46 & 0.90 & 0.10 & 0.00 & 0.36 \\
            DP$^{\dagger}$ & \underline{0.88} & \underline{1.00} & \underline{0.80} & \underline{0.40} & \underline{0.77} \\
            ACT$^{\ddagger}$ & 0.50 & 0.88 & 0.60 & 0.12 & 0.52 \\
            ACT$^{\dagger}$ & 0.64 & 0.90 & 0.46 & 0.04 & 0.51 \\
            $\pi_0$ & 0.66 & 0.86 & 0.48 & 0.22 & 0.55 \\
            GR00T N1.5 & 0.78 & \underline{1.00} & 0.72 & \underline{0.40} & 0.72 \\
            \midrule

            \rowcolor{sacBlue}
            SC-VLA(SPI) & 0.96 & 1.00 & 0.82 & 0.50 & 0.82 \\
            \rowcolor{sacBlue}
            SC-VLA(SPI, OAR) & \textbf{1.00} & \textbf{1.00} & \textbf{0.88} & \textbf{0.56} & \textbf{0.86} \\
            
            $\Delta$ from OAR &
            {\color{deltaGreen}+4\%} &
            {\color{deltaGreen}+0\%} &
            {\color{deltaGreen}+6\%} &
            {\color{deltaGreen}+6\%} &
            {\color{deltaGreen}+4\%} \\
            \bottomrule
        \end{tabular}%
    } 
    \vspace{-0.2cm}
\end{table}

\paragraph{Quantitative Results.}
Table~\ref{tab:success_rates} summarizes the success rates of all methods on four challenging manipulation tasks in ManiSkill. Overall, the proposed \emph{Self-Correcting Vision-Language-Action} method demonstrates superior performance, significantly outperforming existing baseline approaches across all categories. Taking the most challenging \emph{PegInsertion} task as an example, SC-VLA (SPI) improves the success rate by \textbf{28\%} and \textbf{10\%} over pretrained models such as $\pi_0$~\cite{pi_0} and GR00T N1.5~\cite{gr00t}, respectively, demonstrating that explicitly incorporating short-horizon physical state prediction effectively enhances control precision in complex contact scenarios. Building on this, the introduction of residual reinforcement learning further boosts the performance, with SC-VLA (SPI, OAR) achieving an average success rate of 86\%.

In addition, Table~\ref{tab:success_length} reports the average completion length over \textbf{successful episodes} for all methods. SC-VLA attains the shortest average completion length of 157 steps, indicating the highest execution efficiency. This corresponds to a \textbf{43\%} reduction compared to pretrained models such as $\pi_0$~\cite{pi_0}, and a \textbf{8\%} reduction relative to lightweight policies such as Diffusion Policy (DP)~\cite{diffusion}. These results show that the proposed method achieves both more precise execution and higher throughput on complex manipulation tasks.

\begin{table}[t]
    \centering
    \small
    \setlength{\tabcolsep}{5pt} 
    \renewcommand{\arraystretch}{1.15}

    \definecolor{mpgRed}{RGB}{255,235,235}
    \definecolor{sacBlue}{RGB}{235,240,255}
    \definecolor{deltaGreen}{RGB}{0,150,0}
    \definecolor{avgGray}{RGB}{245,245,245}

    \caption{Average completion length over successful episodes.}
    \label{tab:success_length}

    \resizebox{\linewidth}{!}{%
        \begin{tabular}{lcccc>{\columncolor{avgGray}}c}
            \toprule
            \multirow{2}{*}{\textbf{Model}} &
            \multicolumn{4}{c}{\textbf{ManiSkill Tasks}} &
            \multicolumn{1}{c}{\multirow{2}{*}{\textbf{Avg}}} \\
            \cmidrule(lr){2-5}
            & \textbf{\makecell{Stack\\Cube}} 
            & \textbf{\makecell{Place\\Sphere}} 
            & \textbf{\makecell{LiftPeg\\Upright}}  
            & \textbf{\makecell{Peg\\Insertion}}
            & \multicolumn{1}{c}{} \\
            \midrule
      
            DP$^{\ddagger}$ & 162 & 146 & 202 & 800 & 327 \\
            DP$^{\dagger}$ & \textbf{132} & 125 & \underline{197} & 233 & \underline{172} \\
            ACT$^{\ddagger}$ & \underline{148} & 131 & 207 & 250 & 184 \\
            ACT$^{\dagger}$ & 198 & 126 & 203 & \underline{230} & 189 \\
            $\pi_0$ & 265 & 179 & 331 & 331 & 276 \\
            GR00T N1.5 & 192 & \underline{122} & 209 & 257 & 195 \\
            \midrule

            \rowcolor{mpgRed}
            SC-VLA(SPI) & 169 & 128 & 190 & 262 & 187 \\
            \rowcolor{mpgRed}
            SC-VLA(SPI,OAR) & 158 & \textbf{110} & \textbf{189} & \textbf{173} & \textbf{157} \\
            
            $\Delta$ from OAR &
            {\color{deltaGreen}$\downarrow 6.2\%$} &
            {\color{deltaGreen}$\downarrow 14.0\%$} &
            {\color{deltaGreen}$\downarrow 0.5\%$} &
            {\color{deltaGreen}$\downarrow 34.0\%$} &
            {\color{deltaGreen}$\downarrow 16.0\%$} \\
            \bottomrule
        \end{tabular}
    } 
    \vspace{-0.2cm}
\end{table}

\subsection{Ablation Study}
\label{sec:ablation}
To analyze the contribution of each key component in the SC-VLA framework, we conduct a systematic ablation study on the ManiSkill benchmark. All ablation variants are trained using the same demonstration data and evaluated under identical protocols, with the average success rate across the four tasks serving as the primary metric to ensure comparability and fairness.

\begin{table}[htbp]
    \centering
    \small
    \setlength{\tabcolsep}{5pt}
    \renewcommand{\arraystretch}{1.15}

    \caption{Ablation study on different imagination components.}
    \label{tab:ablation_state1}

    \resizebox{\linewidth}{!}{%
        \begin{tabular}{lccccc}
            \toprule
            \multirow{2}{*}{\textbf{Model}} &
            \multicolumn{4}{c}{\textbf{ManiSkill Tasks}} &
            \multirow{2}{*}{\textbf{Avg}} \\
            \cmidrule(lr){2-5}
            & \textbf{\makecell{Stack\\Cube}} 
            & \textbf{\makecell{Place\\Sphere}} 
            & \textbf{\makecell{LiftPeg\\Upright}} 
            & \textbf{\makecell{Peg\\Insertion}}
            & \\
            \midrule

            SC-VLA w/o state       & 0.88 & 1.00 & \textbf{0.84} & 0.42 & 0.78 \\
            SC-VLA w/o prog        & 0.92 & 1.00 & 0.80 & 0.50 & 0.80 \\
            SC-VLA w/o state\_prog & 0.78 & 1.00 & 0.72 & 0.40 & 0.72 \\

            \midrule
            \rowcolor{gray!15} 
            SC-VLA(SPI) & \textbf{0.96} & \textbf{1.00} & 0.82 & \textbf{0.50} & \textbf{0.82} \\

            \bottomrule
        \end{tabular}%
    }
    \vspace{-0.2cm}
\end{table}

\begin{figure*}[t]
    \centering
    \includegraphics[width=0.95\textwidth]{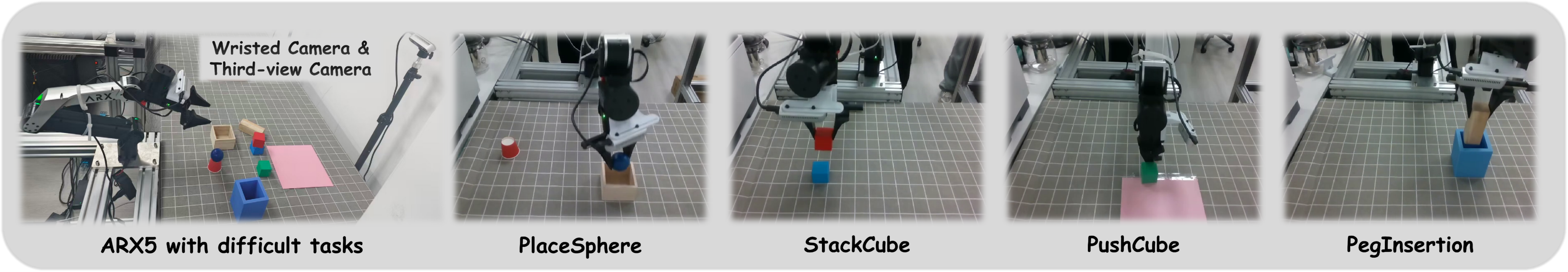}
    \caption{Hardware platforms and visualizations of sampled tasks. The setup is equipped with a wristed camera and a third-person camera.}
    \label{fig:real_world_scene}
    \vspace{-0.2cm}
\end{figure*}

\paragraph{Effectiveness of Progress Guidance.}
The progress token ($\mathbf{q}_{p_t}$) provides explicit temporal progression information for action generation. To evaluate its effectiveness, we conduct an ablation study by removing $\mathbf{q}_{p_t}$ from the query sequence. As shown in Table~\ref{tab:ablation_state1}, removing $\mathbf{q}_{p_t}$ reduces the average success rate from 82\% to 80\%, with the degradation mainly observed on tasks with clear execution stages, such as \emph{StackCube} (96\% $\rightarrow$ 92\%) and \emph{LiftPegUpright} (82\% $\rightarrow$ 80\%). These results indicate that the absence of $q_{p_t}$ does not significantly compromise action reliability, but plays an important role in improving the temporal consistency of the policy during multi-stage execution.

\paragraph{Effectiveness of State Guidance.}
Relative state change modeling ($\Delta s_t$) provides directional constraints on short-term physical evolution for action generation. To evaluate its effectiveness, we conduct an ablation study by removing $\mathbf{q}_{\Delta s_t}$ from the query sequence. As shown in Table~\ref{tab:ablation_state1}, removing $\mathbf{q}_{\Delta s_t}$ leads to a clear drop in the average success rate from 82\% to 78\%, with the most pronounced degradation observed in difficult tasks such as \emph{PegInsertion} (50\% $\rightarrow$ 42\%) and \emph{StackCube} (96\% $\rightarrow$ 88\%). These results indicate that \textit{delta\_state} plays a critical role in stabilizing the physical consistency of action execution, particularly in tasks that are sensitive to contact dynamics and pose accuracy.

\paragraph{Complementarity Between Progress and State Guidance.}
We further examine the interaction between progress guidance ($p_t$) and relative state change modeling ($\Delta s_t$) by jointly removing both components. As shown in Table~\ref{tab:ablation_state1}, eliminating $\mathbf{q}_{p_t}$ and $\mathbf{q}_{\Delta s_t}$ simultaneously causes the average success rate to drop significantly to $72\%$, which is markedly lower than the performance under either single ablation setting. This result demonstrates that progress guidance and state guidance play complementary roles in the policy.
\begin{figure}[h]
    \centering
    \includegraphics[width=0.95\columnwidth]{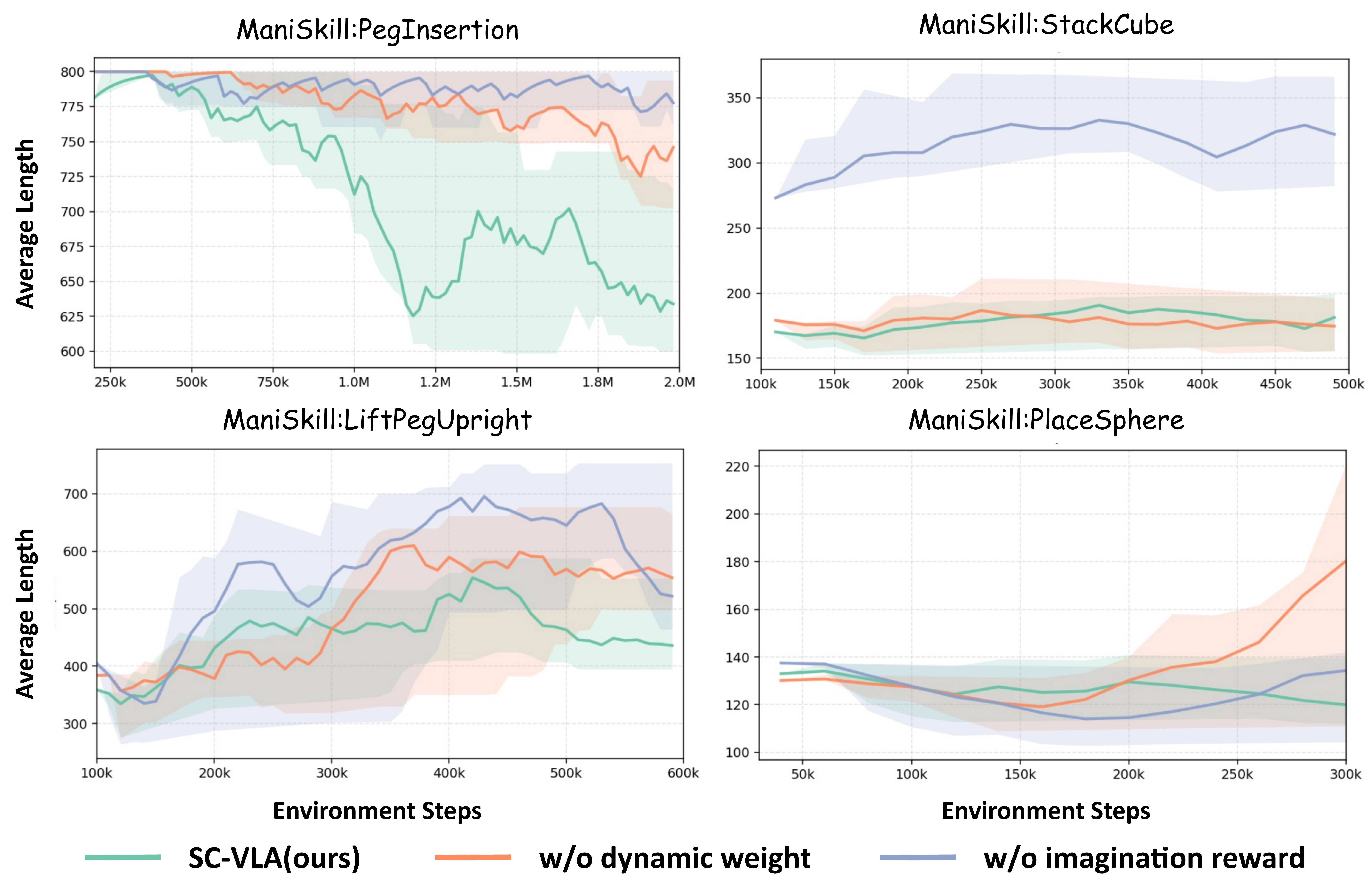}
    \caption{\textbf{Ablation study on the effectiveness of sparse world imagination rewards and dynamic weight scheduling.} We visualize the performance curves starting from the main training phase, excluding the data collection and residual warm-up periods. See Appendix C for further details.}
    \label{fig:ablation_stage2}
    \vspace{-0.2cm}
\end{figure}
\paragraph{Effectiveness of Sparse World Imagination Rewards.}
The sparse world imagination reward plays a pivotal guiding role in reinforcement learning exploration by transforming physical evolution predictions into dense directional feedback. To evaluate its effectiveness, we conducted an ablation study by removing guiding reward and training the residual policy relying solely on the environment's sparse reward. As shown in Fig.~\ref{fig:ablation_stage2}, while the performance gap is negligible in the simple \textit{PlaceSphere} task due to the strong base policy, the reward becomes critical in complex tasks like \textit{PegInsertion}. It provides dense feedback to break exploration bottlenecks, reducing average steps from 800 to 650 and boosting throughput. This confirms its essential role in overcoming cold-start problems for complex manipulation.

\paragraph{Effectiveness of Dynamic Weight Scheduling.}
The dynamic weighting mechanism is responsible for regulating the intensity of prior guidance based on task progress to balance early exploration guidance with later autonomous fine-tuning. To evaluate its effectiveness, we conducted a comparative ablation study by replacing the dynamic decay weight with a fixed constant weight. Results on precision tasks like \textit{PlaceSphere} and \textit{PegInsertion} show that fixed weights lead to significant late-stage degradation, manifesting as divergent step counts or stagnation at sub-optimal solutions. This highlights the conflict between static priors and fine-grained control, demonstrating that progress-based regulation is essential to prevent prior bias from interfering with fine manipulation.

\subsection{Real World Experiments}
\paragraph{Setup.}
We evaluate SC-VLA (SPI) on a real ARX5 arm across four tasks: \emph{StackCube}, \emph{PlaceSphere}, \emph{PegInsertion}, and \emph{PushCube}. Models are trained on 60 demonstrations per task. Hardware setups are visualized in Fig.~\ref{fig:real_world_scene}. Given the difficulty of real-world reward design, we benchmark against widely adopted baselines: GR00T N1.5~\cite{gr00t} and Diffusion Policy (DP)~\cite{diffusion}. We conduct 20 execution trials per task.

\begin{table}[htbp]
    \centering
    \small
    \setlength{\tabcolsep}{5pt}
    \renewcommand{\arraystretch}{1.15}

    \caption{Success rates on real-world tasks using the ARX5 arm.}
    \label{tab:real_bot_res}

    \resizebox{\linewidth}{!}{%
        \begin{tabular}{lccccc}
            \toprule
            \multirow{2}{*}{\textbf{Model}} &
            \multicolumn{4}{c}{\textbf{Real World Tasks}} &
            \multirow{2}{*}{\textbf{Avg}} \\
            \cmidrule(lr){2-5}
            & \textbf{\makecell{Stack\\Cube}} 
            & \textbf{\makecell{Place\\Sphere}} 
            & \textbf{\makecell{Push\\Cube}} 
            & \textbf{\makecell{Peg\\Insertion}}
            & \\
            \midrule

            DP$^{\ddagger}$              & 0.30 & 0.40 & 0.45 & 0.00 & 0.28 \\
            GR00T N1.5      & 0.75 & 0.45 & 0.80 & 0.30 & 0.57 \\

            \midrule
            \rowcolor{gray!15} 
            SC-VLA(SPI)         & \textbf{0.85} & \textbf{0.60} & \textbf{1.00} & \textbf{0.40} & \textbf{0.71} \\

            \bottomrule
        \end{tabular}%
    }
    \vspace{-0.1cm}
\end{table}

\paragraph{Quantitative Results.}
As detailed in Table~\ref{tab:real_bot_res}, SC-VLA (SPI) demonstrates superior capability, achieving an average success rate of 70\% that surpasses DP and GR00T N1.5 by margins of \textbf{43\%} and \textbf{14\%}, respectively.  Its superior performance in contact-rich tasks, such as PegInsertion and StackCube, validates that the sparse world imagination effectively enhances robustness and generalization in complex real-world dynamics.

\section{Conclusion}
We propose \emph{Self-Correcting Vision-Language-Action} (SC-VLA), a framework coupling prediction and control via sparse world imagination. This approach addresses the limitations of static priors and insufficient physical modeling in existing VLA. Our results demonstrate that structured predictive priors can guide physically consistent actions without the complexity of explicit world models. By unifying world-aware modeling with intrinsic reinforcement learning, SC-VLA eliminates the need for manual reward engineering. Extensive experiments confirm that this self-correcting paradigm substantially improves task throughput, offering a robust direction for developing autonomous, self-evolving robotic systems.

\section*{Impact Statement}
This paper presents work whose goal is to advance the field of Machine
Learning. There are many potential societal consequences of our work, none
which we feel must be specifically highlighted here.

\nocite{langley00}

\bibliography{reference}
\bibliographystyle{icml2025}

\newpage
\appendix
\onecolumn

\section*{Appendix A: Task Setup and Evaluation Details}
\label{app:task_details}

This appendix provides supplementary details for the ManiSkill3~\cite{maniskill3} simulation tasks used in the main paper. All tasks are built upon the SAPIEN~\cite{sapien} physics engine and are designed to evaluate robotic manipulation capabilities under high-precision control requirements and complex contact conditions.

\subsection*{A.1 Task Descriptions}

\paragraph{(1) StackCube.}
Stack the cube on top of the other cube: This task requires the robot to grasp a cube and stably stack it on top of another cube. It primarily evaluates pose accuracy during the grasping and placement phases, as well as execution stability. The task is particularly sensitive to positional errors and end-effector jitter.

\paragraph{(2) PlaceSphere.}
Pick up the ball and place it in the target position: In this task, the robot is required to grasp a spherical object and place it at a designated target location. Due to the tendency of the sphere to roll during contact, this task imposes stringent requirements on end-effector velocity control and dynamic stability.

\paragraph{(3) LiftPegUpright.}
Pick up the peg and place it upright: This task requires the robot to grasp one side of a peg and adjust the contact relationship to lift it from a horizontal configuration to an upright pose. During manipulation, relative sliding occurs between the peg and the end-effector. The upright phase demands accurate angle control and pose adjustment, making this task particularly challenging in terms of contact modeling and fine-grained control.

\paragraph{(4) PegInsertion.}
Pick up the peg and insert it into the container next to the peg: This task requires the robot to precisely insert a peg into a hole with small tolerance. It emphasizes accurate pose alignment and fine-grained adjustments during the contact phase, and serves as a standard benchmark for evaluating high-precision manipulation capabilities.

\subsection*{A.2 Training and Evaluation Settings}

For all simulation tasks, all methods are trained using the same amount of demonstration data to ensure a fair comparison. After training, each method is independently evaluated over 50 episodes per task, with task success rate reported as the primary performance metric. Reward definitions follow the official ManiSkill3~\cite{maniskill3} specifications, and the detailed success criteria for each task are summarized in Table~\ref{tab:tasks_success}.

\begin{table}[!h]
    \centering
    \small
    \setlength{\tabcolsep}{5pt}
    \renewcommand{\arraystretch}{1.5}

    \caption{Detailed definitions and success criteria for the four manipulation tasks used in our evaluation.}
    \label{tab:tasks_success}

    \renewcommand{\tabularxcolumn}[1]{m{#1}}

    \begin{tabularx}{\linewidth}{>{\centering\arraybackslash}m{0.16\linewidth} X c}
        \toprule
        \textbf{Tasks} & \multicolumn{1}{c}{\textbf{Success Criteria}} & \textbf{Max Steps} \\
        \midrule
        
        StackCube & 
        The task is considered successful if the horizontal offset of the red cube relative to the green cube does not exceed the half-diagonal of the cube plus 5~mm, the vertical distance between the centers of the two cubes equals one cube edge length with a tolerance of 5~mm, and the red cube remains stationary and is not grasped by the robot end-effector. 
        & 800 \\
        
        PlaceSphere & 
        The task is considered successful if the horizontal offset of the sphere relative to the center of the container does not exceed 5~mm, the vertical position satisfies that the contact point between the bottom of the sphere and the container floor has an error within 5~mm, and the sphere is not grasped by the robot end-effector. 
        & 500 \\
        
        LiftPegUpright & 
        The task is considered successful if the peg’s orientation is close to the upright configuration, with a deviation around the vertical axis not exceeding 0.08~rad, and the vertical position of the peg’s center of mass deviates from its half-length by no more than 5~mm. 
        & 800 \\
        
        PegInsertionside & 
        The task is considered successful if the tail end of the peg lies within the coordinate frame of the box hole, with an offset along the insertion direction no greater than 2~cm, and offsets along the other two orthogonal directions both smaller than 8~mm. 
        & 800 \\
        
        \bottomrule
    \end{tabularx}
\end{table}

\section*{Appendix B: Baseline Details and Settings}
\label{app:baselines}

This appendix provides brief descriptions of the baseline methods used in the main paper. All baselines are trained and evaluated under the same demonstration data budget and evaluation protocols to ensure a fair comparison.

\subsection*{B.1 DP (Diffusion Policy)}
Diffusion Policy~\cite{diffusion} is an imitation learning approach based on diffusion models, which represents continuous action distributions through a progressive denoising process. By effectively modeling multimodal action behaviors, Diffusion Policy has been widely adopted as a strong baseline for robotic manipulation tasks. For the DP baseline, we utilize the implementation provided by the LeRobot~\cite{lerobot} codebase. We configure the policy with an observation history of 2 frames ($n_{\text{obs\_steps}}=2$), a prediction horizon of 16 ($H=16$), and an execution chunk size of 8 ($n_{\text{action\_steps}}=8$). All models are trained for 200,000 iterations on a single NVIDIA RTX 5090 GPU with a batch size of 64. We apply random cropping with a 90\% ratio for data augmentation. All other parameters are set to their default values.

\subsection*{B.2 ACT (Action Chunking with Transformers)}
ACT~\cite{act} combines Conditional Variational Autoencoders (CVAEs) with Transformer architectures to model action sequences in a chunked manner. This design enables the generation of coherent and smooth action trajectories over long horizons, and has demonstrated stable performance in fine-grained manipulation scenarios. For the ACT baseline, we utilize the implementation provided by the LeRobot~\cite{lerobot} codebase. We configure the policy with an observation history of 1 frame ($n_{\text{obs\_steps}}=1$), a prediction horizon of 50 ($H=50$), and an execution chunk size of 20 ($n_{\text{action\_steps}}=20$). All models are trained for 200,000 iterations on a single NVIDIA RTX 5090 GPU with a batch size of 64. All other parameters are set to their default values.

\subsection*{B.3 $\boldsymbol{\pi}_0$}
$\pi_0$~\cite{pi_0} is a recently proposed robot foundation model developed by Physical Intelligence, which acquires broad robotic manipulation capabilities through large-scale, multi-task and cross-embodiment training. Built upon a pre-trained vision-language model and augmented with a flow-based action generation module, $\pi_0$~\cite{pi_0} demonstrates strong performance on a wide range of real-world robotic manipulation tasks, including long-horizon and highly dexterous scenarios. For the $\pi_0$ baseline, we utilize the official implementation. The model is trained for 50,000 iterations on a single NVIDIA RTX PRO 6000 GPU with a batch size of 32. We employ a cosine decay learning rate schedule, configured with 3,000 warmup steps, a peak learning rate of $2 \times 10^{-5}$, and a final learning rate of $1.5 \times 10^{-6}$ (decaying over 100,000 steps). All other parameters are set to their default values.

\subsection*{B.4 GR00T N1.5}
GR00T N1.5~\cite{gr00t} is a robot manipulation model based on flow matching, which uses the DiT architecture to model continuous action distributions and achieves efficient and stable action generation through vector field learning. This paper extends this method based on its underlying architecture and uses it as a base model. For the GR00T N1.5 baseline, we utilize the official implementation. The model is trained for 50,000 iterations on a single NVIDIA L40 GPU with a batch size of 32. The learning rate is set to $1 \times 10^{-4}$. All other parameters are set to their default values.

\section*{Appendix C: Experiment Details}
\label{appendix:training_phases}

\paragraph{Multi-Stage Training Protocol.}
To ensure the stability of the hybrid architecture, integrating a pre-trained Flow Matching DiT base policy with a randomly initialized SAC residual module, we implement a three-stage training protocol. As illustrated in Fig.~\ref{fig:prog_explore}, the residual weight $\lambda$ is dynamically adjusted to facilitate a smooth transition from pure imitation to residual reinforcement learning:

\begin{figure*}[h]
    \centering
    \includegraphics[width=0.3\textwidth]{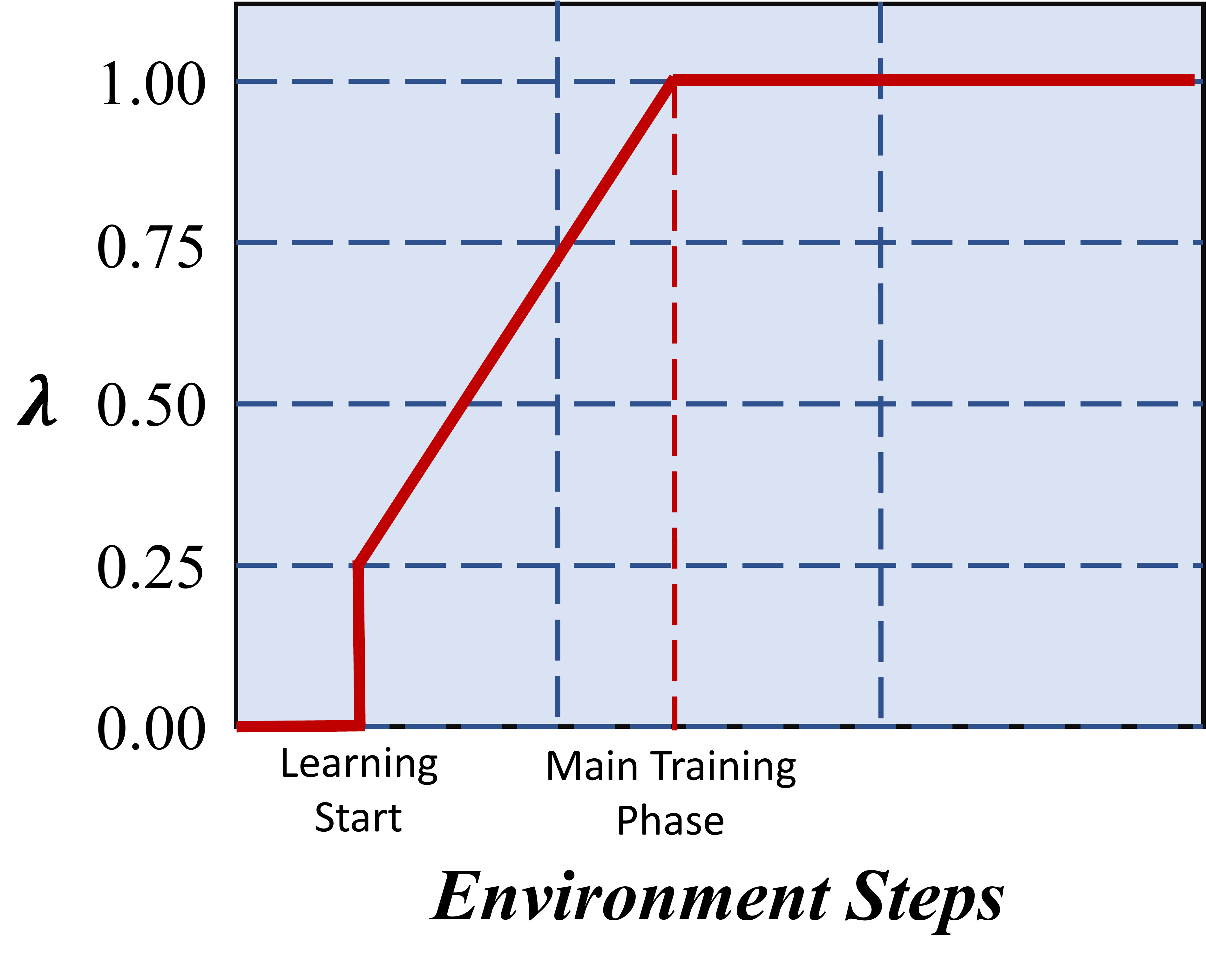}
    \caption{\textbf{Schematic of the multi-stage training protocol and residual weight schedule.} The training process is divided into three distinct phases to ensure stability: (1) \textit{Buffer Warm-up} for gathering demonstrations, (2) \textit{Residual Injection} for gradually introducing the RL policy, and (3) the \textit{Main Training Phase}. The curve illustrates the evolution of the residual weight $\lambda$, transitioning from pure imitation to full residual learning.}
    \label{fig:prog_explore}
\end{figure*}

\begin{enumerate}
    \item \textbf{Initial Data Collection (Buffer Warm-up):} In the very early stage, the residual weight is held constant at $\lambda=0$. The agent acts solely using the frozen base policy to collect high-quality demonstration trajectories. This phase is strictly for populating the SAC replay buffer with valid transitions, ensuring the critic has a meaningful state distribution to learn from before any policy updates occur.
    
    \item \textbf{Residual Injection (Policy Warm-up):} Once the buffer is sufficiently populated, we enter a linear warm-up phase. The residual weight $\lambda$ is linearly increased from 0 to the target scaling factor. This gradual injection prevents the noise from the initially random residual policy from catastrophically disrupting the control loop, allowing the RL agent to adapt to the base policy's dynamics without causing immediate task failure.
    
    \item \textbf{Main Training Phase:} Upon reaching the target weight, $\lambda$ remains fixed (subject to the dynamic weight scheduling described in Sec.~4.2), and the formal training and evaluation begin.
\end{enumerate}

\paragraph{Visualization Rationale.}
It is important to note that the \textit{Data Collection} and \textit{Residual Injection} phases are transient and brief compared to the overall training duration. They serve merely as necessary initialization steps to stabilize the learning environment. During these preliminary phases, performance metrics naturally fluctuate: the agent exhibits stable behavior during data collection (using the base policy) but experiences an inevitable ``adaptation dip'' during residual injection as the RL module explores. Detailed parameter settings for these phases are provided in~\cref{tab:sac_hyperparameter}.
To provide a clear and fair comparison of the proposed methods (\textit{Sparse World Imagination Reward} and \textit{Dynamic Weight Scheduling}), we exclude these initialization artifacts from our results. The performance curves in Fig.~\ref{fig:ablation_stage2} strictly record the \textit{Main Training Phase}, starting immediately after the residual weight reaches its target value. This ensures the visualization focuses on the model's convergence behavior and asymptotic performance rather than its setup dynamics.

\section*{Appendix D: Implementation Details}
\label{app:implementation_details}

To ensure reproducibility, this appendix details the key configurations and hyperparameter settings used in the two training stages of the SC-VLA framework.

\paragraph{Stage I: SC-VLA Base Policy Training.}
The base policy is implemented using the DiT architecture from \emph{GR00T N1.5}. We jointly optimize the flow-matching objective and the multimodal prediction heads using the AdamW optimizer. The model is trained for 50,000 iterations on a single NVIDIA L40 GPU. The detailed network configurations and training hyperparameters are summarized in Table~\ref{tab:base_hyperparameter}.

\begin{table}[h]
    \centering
    \small
    \caption{SC-VLA base model hyperparameter settings.}
    \label{tab:base_hyperparameter}

    \begin{tabularx}{0.6\textwidth}
    {>{\centering\arraybackslash}X >{\centering\arraybackslash}m{0.3\textwidth}}
        \toprule
        \textbf{Hyperparameter} & \textbf{Value} \\
        \midrule
        Batch size       & 32       \\
        Training steps   & 50{,}000 \\
        Learning rate    & $1 \times 10^{-4}$     \\
        Model init seed  & 42       \\
        Optimizer        & AdamW    \\
        \bottomrule
    \end{tabularx}
\end{table}

\paragraph{Stage II: SC-VLA Residual Policy Training.}
In the second stage, the residual module is trained using the Soft Actor-Critic (SAC) algorithm. During this phase, the parameters of the base model are fully frozen, and optimization is performed exclusively on the residual network $\pi_{\text{res}}$. 

To account for performance variance of the base policy across different tasks, we adopt a task-adaptive residual scaling strategy. Specifically, following an inverse-correlation principle, tasks for which the base model achieves higher success rates are assigned smaller residual scaling and evaluation coefficients, enabling fine-grained refinement while preserving high-quality priors. Conversely, larger coefficients are used for tasks where the base policy performs worse, allowing for more substantial action corrections.

Although the residual scaling factor varies across tasks, we employ a consistent dynamic scheduling strategy for the predictive guidance reward across all experiments. Key hyperparameters used in the reinforcement learning stage are summarized in Table~\ref{tab:sac_hyperparameter}.

 \begin{table}[h]
    \centering
    \small
    \caption{SC-VLA residual model hyperparameter settings.}
    \label{tab:sac_hyperparameter}

    \begin{tabularx}{1\textwidth}{
        >{\centering\arraybackslash}m{0.15\textwidth} 
        >{\raggedright\arraybackslash}m{0.2\textwidth} 
        >{\raggedright\arraybackslash}m{0.2\textwidth} 
        >{\raggedright\arraybackslash}X                   
        >{\raggedright\arraybackslash}m{0.12\textwidth} 
        }
        \toprule
        & \multicolumn{2}{c}{\textbf{Shared Parameters}} & \multicolumn{2}{c}{\textbf{Task-Specific Parameters}} \\
        \cmidrule(lr){2-3}\cmidrule(lr){4-5}
        
        \textbf{Tasks} & 
        \multicolumn{1}{c}{\textbf{Hyperparameter}} & 
        \multicolumn{1}{c}{\textbf{Value}} & 
        \multicolumn{1}{c}{\textbf{Hyperparameter}} & 
        \multicolumn{1}{c}{\textbf{Value}} \\
        \midrule

        \multirow{5}{*}{StackCube} &
        
        \multicolumn{2}{c}{\multirow{20}{*}{
            \renewcommand{\arraystretch}{1.3} 
            \begin{tabular}{@{}p{0.2\textwidth} p{0.2\textwidth}@{}}
                Random seed & 0 \\
                Replay buffer size & Total steps \\ 
                Discount factor $\gamma$ & 0.97 \\
                Target smoothing $\tau$ & 0.01 \\
                Batch size $B$      & 1024 \\
                Policy learning rate & $1 \times 10^{-4}$ \\
                Critic learning rate & $1 \times 10^{-4}$ \\
                Entropy coeff. $\alpha$ & 0.2 \\
                Gradient clipping & 50 \\
                Update-to-Data (UTD) & 0.5 \\
                Training frequency & 64 \\
                Guide weight $w_{\text{guide}}$ & 0.6
            \end{tabular}
        }} &
        
        Total timesteps   & 500,000 \\
        & \multicolumn{2}{c}{} & Learning start steps   & 30,000 \\
        & \multicolumn{2}{c}{} & Exploration steps      & 100,000 \\
        & \multicolumn{2}{c}{} & Residual scale (train) & 0.01 \\
        & \multicolumn{2}{c}{} & Residual scale (eval)  & 0.005 \\
        \cmidrule(lr){4-5}

        \multirow{5}{*}{PlaceSphere}
        & \multicolumn{2}{c}{} & Total timesteps   & 500,000 \\
        & \multicolumn{2}{c}{} & Learning start steps   & 8,000 \\
        & \multicolumn{2}{c}{} & Exploration steps      & 30,000 \\
        & \multicolumn{2}{c}{} & Residual scale (train) & 0.01 \\
        & \multicolumn{2}{c}{} & Residual scale (eval)  & 0.005 \\
        \cmidrule(lr){4-5}

        \multirow{5}{*}{LiftPegUpright}
        & \multicolumn{2}{c}{} & Total timesteps   & 600,000 \\
        & \multicolumn{2}{c}{} & Learning start steps   & 30,000 \\
        & \multicolumn{2}{c}{} & Exploration steps      & 100,000 \\
        & \multicolumn{2}{c}{} & Residual scale (train) & 0.01 \\
        & \multicolumn{2}{c}{} & Residual scale (eval)  & 0.005 \\
        \cmidrule(lr){4-5}

        \multirow{5}{=}{\centering PegInsertion}
        & \multicolumn{2}{c}{} & Total timesteps   & 3,000,000 \\
        & \multicolumn{2}{c}{} & Learning start steps   & 60,000 \\
        & \multicolumn{2}{c}{} & Exploration steps      & 200,000 \\
        & \multicolumn{2}{c}{} & Residual scale (train) & 0.1 \\
        & \multicolumn{2}{c}{} & Residual scale (eval)  & 0.03 \\

        \bottomrule
    \end{tabularx}
\end{table}

 \begin{figure*}[h]
    \centering
    \includegraphics[width=0.95\textwidth]{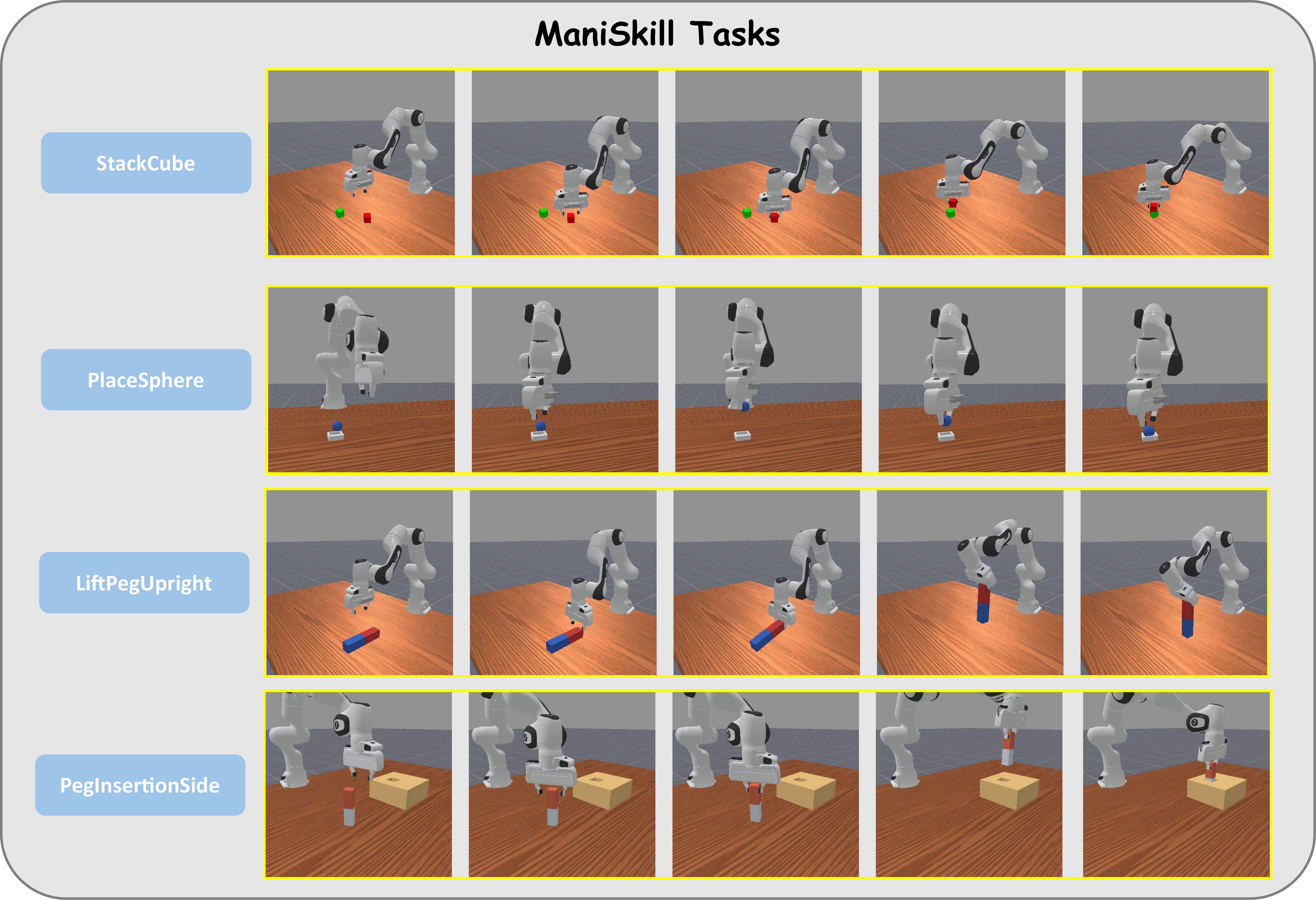}
    \caption{Visualization results of Maniskill simulated tasks.}
    \label{fig:simulated_comic}
\end{figure*}

\begin{figure*}[h]
    \centering
    \includegraphics[width=0.95\textwidth]{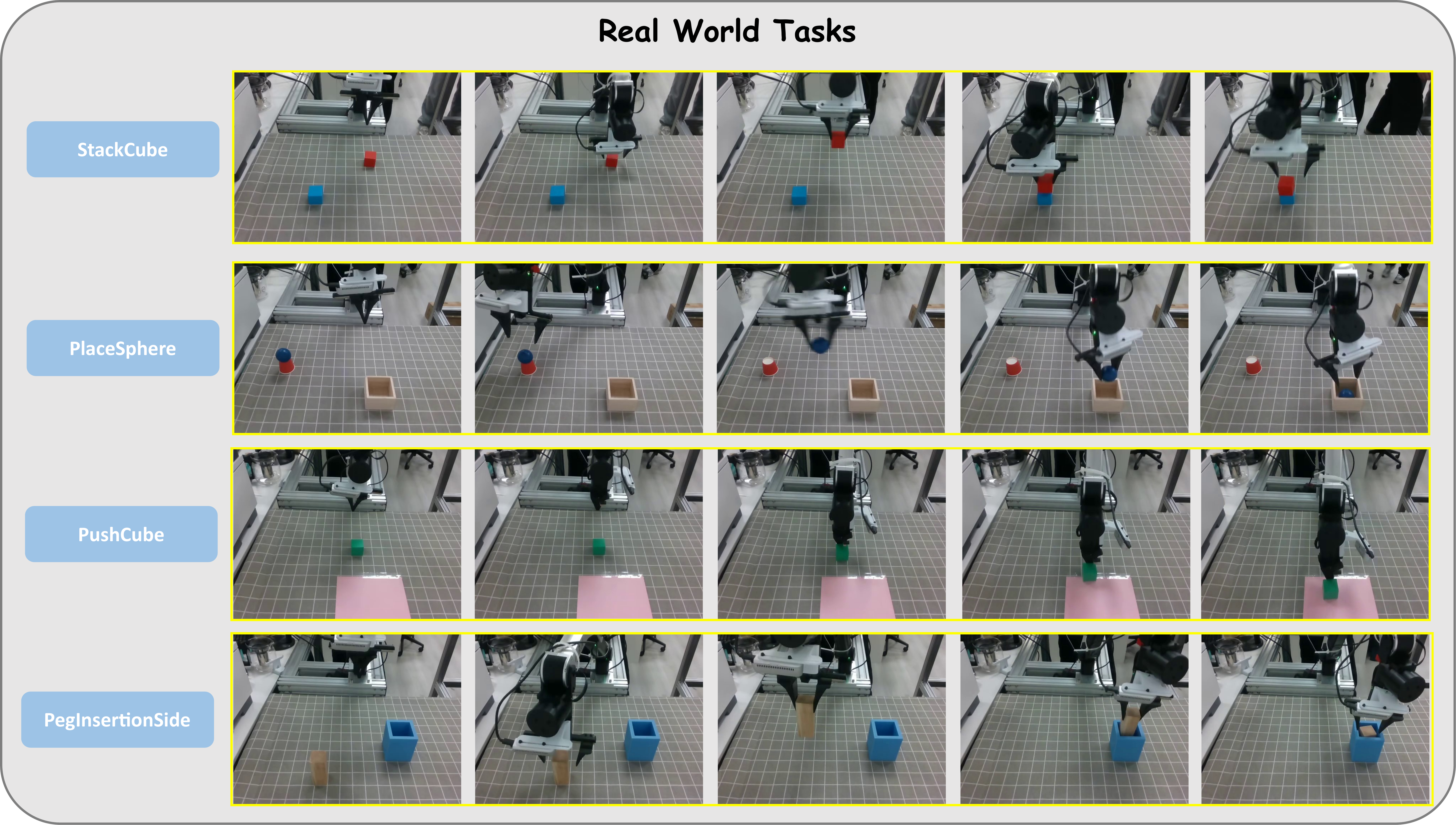}
    \caption{Visualization results of real-world tasks.}
    \label{fig:real_world_comic}
\end{figure*}


\end{document}